\documentclass[runningheads]{llncs}

\usepackage[final,year=2024,ID=xxx]{eccv}
\usepackage{eccv}
\usepackage{eccvabbrv}
\usepackage{graphicx}
\usepackage{amsmath}
\usepackage{amssymb}
\usepackage{cancel}
\usepackage{booktabs}
\usepackage{xcolor}
\usepackage{enumitem}
\usepackage{nicefrac}
\usepackage{circledsteps}
\usepackage{multirow}
\usepackage{arydshln}
\usepackage{pifont}
\usepackage{nccmath}
\usepackage{lipsum}
\usepackage{subcaption}
\usepackage{booktabs}
\usepackage{amssymb,mathrsfs,amsmath}
\usepackage{wrapfig}
\usepackage{marvosym}

\DeclareMathAlphabet\mathbfcal{OMS}{cmsy}{b}{n}
\newcommand{\yes}{\ding{51}}

\usepackage[pagebackref,breaklinks,colorlinks]{hyperref}
\usepackage[capitalize]{cleveref}
\crefname{section}{Sec.}{Secs.}
\Crefname{section}{Section}{Sections}
\Crefname{table}{Table}{Tables}
\crefname{table}{Tab.}{Tabs.}

\definecolor{purple(html/css)}{rgb}{0.5, 0.0, 0.5}
\definecolor{purple(x11)}{rgb}{0.63, 0.36, 0.94}

\begin{document}

\title{CountFormer: Multi-View Crowd Counting Transformer}

\author{
\textsuperscript{1}Hong Mo\textsuperscript{\Letter}$^\dag$  \and
\textsuperscript{2}Xiong Zhang$^\dag$ \and
\textsuperscript{3}Jianchao Tan \and
\textsuperscript{2}Cheng Yang \\ \and
\textsuperscript{1}Qiong Gu \and
\textsuperscript{1}Bo Hang~ \and
\textsuperscript{4}Wenqi Ren
}
\authorrunning{{\bf CountFormer} {H. Mo \it et al.}}
\institute{
  $^1~$Hubei University of Arts \& Science,~~ $^2~$Neolix Autonomous Vehicle \\
  $^3~$Kuaishou Technology,~~~ $^4~$Sun Yat-Sen University \\
  {\footnotesize $\dag$~{Equal contributions}~~ {\Letter}~Corresponding author \href{mailto:mandymo@hbuas.edu.cn}{mandymo@hbuas.edu.cn}}
}

\maketitle
\def\thefootnote{}\footnotetext{\scriptsize This Research is Supported by the Open Project Program of State Key Laboratory of Virtual RealityTechnology and Systems, Beihang University (NO.VRLAB2024C05)}
\vspace{-10pt}
\begin{abstract}
Multi-view counting (MVC) methods have shown their superiority over single-view counterparts, particularly in situations characterized by heavy occlusion and severe perspective distortions. 

However, hand-crafted heuristic features and identical camera layout requirements in conventional MVC methods limit their applicability and scalability in real-world scenarios.
In this work, we propose a concise 3D MVC framework called \textbf{CountFormer}
to elevate multi-view image-level features to a scene-level volume representation and estimate the 3D density map based on the volume features. 
By incorporating a camera encoding strategy, CountFormer successfully embeds camera parameters into the volume query and image-level features, enabling it to handle various camera layouts with significant differences.
Furthermore, we introduce a feature lifting module capitalized on the attention mechanism to transform image-level features into a 3D volume representation for each camera view. 
Subsequently, the multi-view volume aggregation module attentively aggregates various multi-view volumes to create a comprehensive scene-level volume representation, allowing CountFormer to handle images captured by arbitrary dynamic camera layouts. 
The proposed method performs favorably against the state-of-the-art approaches across various widely used datasets, demonstrating its greater suitability for real-world deployment compared to conventional MVC frameworks.
\end{abstract}

\vspace{-15pt}
\section{Introduction}
\vspace{-5pt}
Single-view counting (SVC) has exhibited promising effectiveness, yielding remarkable achievements on well-established datasets \cite{zhang2016single,zeng2017multi,babu2017switching,li2018csrnet,cao2018scale,liu2019crowd,liu2019adcrowdnet,mo2020background,wang2020distribution,song2021rethinking,lian2021locating,mo2022attention,cheng2022rethinking,ranasinghe2023diffuse,wei2023semi}, while the inherent limitations of existing SVC approaches hinder their practical application, thereby impeding their effective deployment in real-world scenarios.

Recently, there has been a growing trend towards addressing the multi-view counting (MVC) problem \cite{zhang2019wide,zhang2022wide,zhang20203d,zhang20223d,zhang2021cross,zhang2022calibration,zhai2022co}.
Specifically, existing approaches in the field of MVC typically make use of flat ground-plane assumptions 
to transform the image-level features onto the ground plane.
Subsequently, fusing the multi-view (MV) ground features to recover scene-level features, 
and estimating the scene-level density using predictors based on fused ground features.

However, the flat ground assumptions \cite{zhang2019wide,zhang2022wide,zhang20203d,zhang20223d,zhang2021cross,zhang2022calibration,zheng2021learning} are not always guaranteed, which can result in misalignment between the scene-level features and the real-world environment, leading to less accurate counting performance.
Additionally, in the MV feature fusion module, the attention weights are determined solely based on the distance from the ground plane to each camera without considering the features themselves \cite{zhang2019wide,zhang2022wide,zhang2021cross,zhang2022calibration,zheng2021learning}, which may limit the effectiveness of the fusion strategy.
Moreover, most current approaches\cite{zhang2019wide,zhang2022wide,zhang20203d,zhang20223d,zhang2022calibration,zheng2021learning} can only handle statically fixed camera layouts, i.e., the camera configurations are identical during the training and inference stage, lacking the ability to perform the MVC task with images from arbitrary dynamic camera settings.

\textcolor{black}{Simultaneously, noteworthy advancements have been accomplished in the domain of multi-view perception (MVP).
Specifically, research works \cite{philion2020lift,hu2021fiery,li2022bevformer,huang2021bevdet,li2023fb,liao2022maptr,li2023lanesegnet} focus on lifting image-level features of MV images to the cohesive scene-level space, followed by the execution of specific perception tasks using the derived scene-level features.
Despite the concise architecture design and the promising performances on the MVP task, it is infeasible to naively adopt existing MVP methods without considering the specific challenges of the MVC settings because contemporary MVP approaches necessitate the assumption of stable and fixed camera layouts, which may not hold in MVC scenarios.
Moreover, the inherent philosophies of employing multi-camera layouts in MVC and MVP are different.
The MVP tasks require multi-view cameras with limited overlapped field of view (FOV) to provide a 360$^\circ$ FOV of the scene, while MVC tasks rely on multi-view settings that the FOV of each camera has significant overlap with the others to address occlusion and scale variation challenges.
}

In this work, we incorporate the recent advanced ideas in MVP and propose an innovative MV learning framework called CountFormer to extend the applicability and scalability of the existing MVC approaches.
CountFormer is founded upon a primary consideration, where it shall be adequate to process images captured with arbitrary dynamic camera layout settings and robust enough to alleviate the performance drop caused by extrinsic parameter perturbation.
To accomplish this objective, a feature lifting module is first proposed to lift the image-level feature in 3D space for each view. 
Specifically, the 3D scene is voxelized into individual voxels, and features corresponding to each voxel and camera view are obtained using a deformable attention mechanism \cite{zhu2020deformable}. Since the feature lifting module does not necessitate the flat ground assumption, the CountFormer is more suitable to deal with challenging situations such as congested crowds or uneven terrain.
In addition, the integration of the attention mechanism enhances its robustness against fluctuations in camera extrinsic parameters, which is particularly beneficial in practical settings where such perturbations are inevitable \cite{yuan2020rggnet}.

Subsequently, an MV volume aggregation module is introduced to attentively fuse the MV volume features to generate the comprehensive scene-level volume representation, where the blending weights are estimated by implicitly joint considering the voxel features and the geometry property of the cameras, ensuring that both the visual features and the geometric information are considered during the fusion process, leading to more accurate and robust results than previous methods \cite{zhang2019wide,zhang2022wide,zhang2022calibration,zheng2021learning,zhang2021cross}.
Due to the sophisticated design, the aggregation module is capable of efficiently handling a dynamic number of volume features, allowing the CountFormer to adapt to scenarios where the number of cameras may vary, ensuring robust performance in various applications.
Afterwards, CountFormer employs 3D convolution operators to estimate the 3D scene-level density map from the aggregated volume representation.

\begin{figure*}[t]
  \centering
  \includegraphics[width=0.9\textwidth]{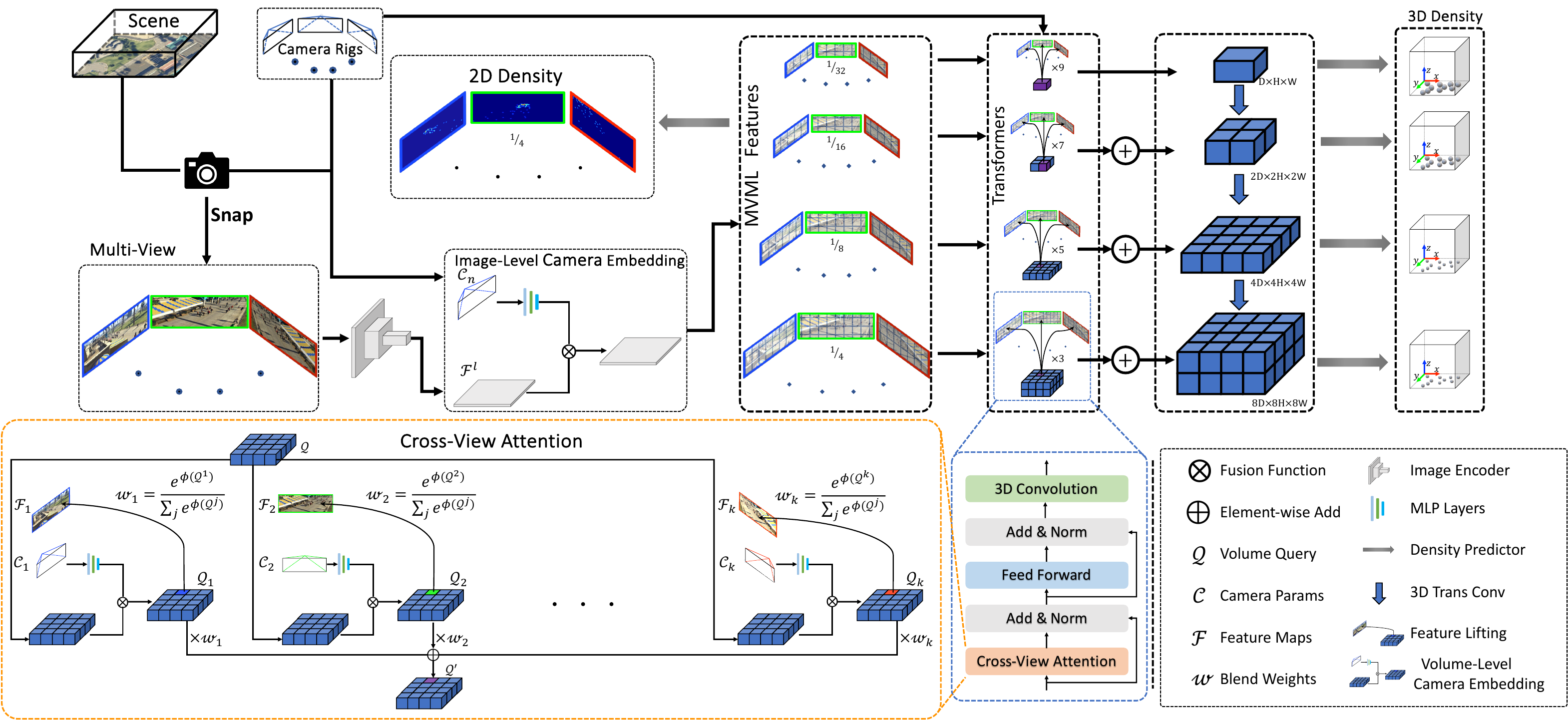}
  \vspace{-7pt}
  \caption{{\bf Framework of the CountFormer.} 
  The {Image Encoder} extracts multi-view and multi-level features (MVML) from the multi-view images of the scene. 
  {Image-Level Camera Embedding Module} fuses camera intrinsic and extrinsic with the MVML features.
  The elaborate {Cross-View Attention Module}, a sophisticated attention component, transforms the image-level features into scene-level volume representations.
  Besides main components, a 2D Density Predictor is used to estimate the image space density, 3D Density Predictors are employed to regress for the 3D scene-level density, and a simple feature pyramid network fuses the multi-scale voxel features.
  }
  \label{fig:arch}
  \vspace{-10pt}
\end{figure*}

To enhance the robustness and representation capability of CountFormer in handling arbitrary dynamic camera layouts, 
the camera encoding strategy is incorporated, which involves implicitly encoding the extrinsic and intrinsic camera parameters. 
Specifically, the camera information is implicitly encoded into the volume query that is used to lift the corresponding image-level features in 3D space. 
Additionally, the camera information is also encoded into the corresponding image-level features. 
By incorporating the camera information into both the volume query and image-level features, CountFormer effectively integrates the camera-specific characteristics into its processing pipeline, leading to improved performance in various camera configurations and scene understanding tasks.

In summary, CountFormer is highly versatile and scalable, making it well-suited and efficient for usage in real-world scenarios, and our main contributions are as follows:
\begin{enumerate}[label=\textbullet]
  \setlength{\itemsep}{0pt}
  \setlength{\parskip}{3pt}
  \vspace{-3pt}
  \item We creatively design a revolutionary multi-view counting (MVC) framework, called CountFormer, which is the first attempt to solve the 3D MVC problem to fit a real-world environment.
  \item A feature lifting module and an MV volume aggregation module are conceived to transform the MV image-level features w.r.t arbitrary dynamic camera layouts into a unified scene-level volume representation.
  \item We present an effective strategy to embed the camera parameters into the image-level features and the volume query, facilitating accurate and adaptable representation among diverse camera setups.
\end{enumerate}

\section{Related Work}
\vspace{-4pt}
Due to the extensive scope of related works, 
we only discuss works that are strongly related to our framework.

{\bf Single-View Counting (SVC).}
Since the work of \cite{zhang2015cross}, density estimation has been the main paradigm for crowd counting.
Subsequent works, such as \cite{ma2022fusioncount,zhang2018crowd,ma2019bayesian,shi2020real,liu2020weighing,zhang2016single,boominathan2016crowdnet,babu2017switching,sindagi2017cnn,liu2018decidenet,shi2019revisiting,jiang2020attention,gao2019pcc,lei2021towards,liu2023point} continue improving performance by designing more powerful model structures that capable of learning multi-scale representation and perspective-free feature detection.
Simultaneously, another line of works \cite{hu2020count,mo2020background,li2018csrnet,du2023redesigning,wei2023semi,cao2018scale,wang2020mobilecount,song2021choose,zhang2019attentional,liu2019context} adopt the insight from semantic image segmentation to exploit the encoder-decoder architecture to facilitate the multi-scale learning capability.
Recently, by adopting the transformer architecture\cite{vaswani2017attention,dosovitskiy2020image}, works such as \cite{gao2022congested,tian2021cctrans,sun2021boosting,liang2022end,wei2021scene,liang2022transcrowd,mo2022attention,yang2022crowdformer,fang2020multi,pan2020attention} substantially advance performances because of their remarkable representation and generalization capability.
In addition to the intricate structures, \cite{ma2021towards,liu2022leveraging,yuan2022translation,gao2023forget,huang2023counting,zhu2023daot,ranasinghe2023diffuse} advances the progress by optimizing the training process or improving the cross-scene generalization ability.
Despite the promising performance, SVC methods encounter challenges that necessitate resolution, including effectively dealing with scale variation, mitigating occlusion difficulties, and handling congested crowds.

{\bf Multi-View Perception (MVP).}
The pioneering research \cite{philion2020lift} introduces a novel approach that involves the transformation of multi-view multi-level features into a cohesive scene-level space, followed by the execution of specific perception tasks using the derived scene-level features.
Subsequently, certain studies \cite{hu2021fiery,liang2022bevfusion,liu2023bevfusion,li2023bevdepth} enhance the framework by incorporating sophisticated fundamental components. 
For instance, these studies utilize the transformer structure as the foundational framework \cite{yang2023bevformer}, employ a robust perception head \cite{xu2022cobevt,liao2022maptr}, integrate multiple modal sensors to enhance performance \cite{liu2023bevfusion,liang2022bevfusion}, and leverage the attention mechanism to effectively encode image-level features to the scene-level space \cite{li2022bevformer,zhou2022cross}.
Furthermore, researchers tackle the task of 3D semantic occupancy prediction by employing a voxelization technique to convert the scene into discrete voxels and subsequently conduct 3D semantic segmentation to retrieve semantic features for each voxel \cite{wei2023surroundocc,zhang2023occformer,li2023fb,huang2023tri,wang2023openoccupancy,tong2023scene}.
Despite the solid theoretical foundation and the extensive experimental validation of their effectiveness,
naively transferring existing MVP approaches to solve the MVC task proves to be difficult, 
where the biggest challenge remains that MVC necessitates the capability of handling arbitrary dynamic camera layouts, which is infeasible for existing methods.

{\bf Multi-View Counting (MVC).}
Pioneering works \cite{zhang2019wide,zhang2022wide,zhang2021cross} propose to transform each camera view's image-level features to the scene's ground plane and then fuse these features to estimate the scene-level density map.
To further improve the performance, \cite{zhang20203d,zhang20223d} consider individuals' variable height in the 3D environment by 
introducing multi-height ground planes along the z-axis. 
Although demonstrating promising performance, the strong assumption limits its applicability and scalability.
Specifically, the feature transform module \cite{zhang2019wide,zhang2022wide,zhang2022calibration,zhang2021cross,zhang20203d,zhang20223d} necessitated a flat ground assumption, which is not guaranteed in real-world situations.
Furthermore, the fusion module's blending weights are exclusively determined by the geometric position of the cameras \cite{zhang2019wide,zhang2022wide,zhang2021cross,zhang2022calibration}, ignoring the critical semantic features. However, these approaches seem counterintuitive, as occlusions and other crucial information related to the scene are primarily encoded in the semantic features.
Alternatively, \cite{zhang20203d,zhang20223d} attempt to achieve reasonable fusion results by jointly considering geometric and semantic features, while these approaches come at the expense of limiting the flexibility and adaptability of the methods to handle dynamic camera layout settings. 

\section{Methodology}

\subsection{Image Encoder}
\vspace{-3pt}
The image encoder aims to extract multi-view and multi-level (MVML) features $\left \{ \mathcal{F}_\mathrm{n}^\mathrm{l} \right \}_{\mathrm{n}\le \mathrm{N},~\mathrm{l}\le \mathrm{L}}$  from multiple synchronized captured images $\left \{ \mathcal{I}_\mathrm{n}\right \}_{\mathrm{n}\le \mathrm{N}}$, 
where varying $\mathrm{N}$ is the number of views, 
and $\mathrm{L}$ indicates the maximum feature pyramid level.
Each image $\mathcal{I}_\mathrm{n}$ is first fed into a vision transformer \cite{liu2021swin} to extract multi-level features.
Subsequently, considering the significance of multi-scale feature fusion, the feature pyramid network \cite{lin2017feature} is adopted to aggregate and dispatch the multi-scale features to enrich the multi-scale representation and contextual information encoding capability, which are also employed by extensive research works \cite{ghiasi2019fpn,zhang2021dcnas,li2023bevdepth,huang2021bevdet,kirillov2019panoptic}.

\subsection{Image-Level Camera Embedding}
The camera parameters play a key role in transforming image-level features into scene-level representation.
\textcolor{black}{Existing approaches tend to employ channel-wise camera encoding to differentiate features across different views \cite{huang2021bevdet,li2022bevformer,liu2022petr,li2023lanesegnet,man2023bev,liu2023sparsebev}.}
However, the existing strategy proves inefficient when dealing with the challenging CVCS dataset \cite{zhang2021cross}, which includes tens of thousands of diverse camera configurations.
\textcolor{black}{In this study, we devise both channel-wise and spatial-wise camera encoding to facilitate learning from MV images with significantly varied camera layouts.}
Formally, given the extrinsic parameters ${\mathrm{M}} \in \mathbb{R}^{3\times 4}$ representing transformation from scene space to camera space, 
the intrinsic parameters $\mathrm{K} \in \mathbb{R}^{3\times 3}$, 
and the image-level augmentation matrix $\mathrm{A_c} \in \mathbb{R}^{2\times 3}$,
we flatten, concatenate, and extend the $\{\mathrm{M,K,A_c}\}$ to build the spatial-aligned camera-parameter vector $\xi \in \mathbb{R}^{27\times \mathrm{H}\times \mathrm{W}}$. 
Then we generate the positional encoding $\mathcal{{P}}_{\rm{c}} \in \mathbb{R}^{2\times \mathrm{H}\times \mathrm{W} }$ defined with $\mathcal{P}_{\rm{c}}(\cdot,\mathrm{u,v})=[\nicefrac{\mathrm{u}}{\mathrm{W}}, \nicefrac{\rm{v}}{\mathrm{H}}]^\mathrm{T}$.
Suppose the feature map is $\mathcal{F} \in \mathbb{R}^{\mathrm{C}\times \mathrm{H}\times \mathrm{W}}$, then the image-level camera  embedding is performed with
\begin{align}
  \mathrm{E}_{\rm{c}}
  \Big(\mathcal{F},\{\mathrm{M,K,A_c}\}\Big)=\mathcal{F}\otimes \phi_{\rm{c}}\Big([\xi, \mathcal{P}_{\rm{c}}]\Big)  \label{eq:camera_embed}, 
\end{align}
where $[\cdot]$ denotes tensor concatenation, 
$\phi_{\rm{c}}$$:\mathbb{R}^{29}\rightarrow\mathbb{R}^{\mathrm{C}}$ represents a Multi-Layer Perceptron (MLP) that aggregates the positional encoding and the camera parameters, 
and $\otimes$ remains a binary operator, 
{such as } the Hadamard Product \cite{kim2016hadamard} or the widely-used element-wise addition.
\subsection{Multi Scale Transformers}

The multi-scale transformers aim to generate the multi-scale volume representations $\{\mathcal{V}_1,\mathcal{V}_2,\cdots,\mathcal{V}_\mathrm{L}\}$ from the MVML features $\left \{ \mathcal{F}_\mathrm{n}^\mathrm{l} \right \}_{\rm{n}\le \mathrm{N},l\le \mathrm{L}}$, where $\mathrm{N}$ is the number of employed view and $\mathrm{L}$ refers the maximum feature pyramid level.
To accomplish this objective, for each level $\mathrm{l}$,  
a CountFormer is dedicated to lift the MV image-level features $\{\mathcal{F}^\mathrm{l}_\mathrm{n}\}_{\mathrm{n} \le \mathrm{N}}$ to the scene-level volume representation $\mathcal{V}_\mathrm{l}$.

Specifically, the CountFormer consists of multiple encoder layers, 
each following the standard design principles of transformers \cite{vaswani2017attention}, 
with three specialized components, 
i.e., the learnable volume query, the cross-view attention, and a computationally efficient 3D convolution that replaces the self-attention mechanism.
To elaborate further, the volume query is used as the initial query for each camera view, which is then combined with camera extrinsic and intrinsic parameters to create the view-dependent query.
The primary purpose of the cross-view attention module is to lift the image-level features to volume features for each camera view and then aggregate the MV volumes to produce a comprehensive scene-level volume representation.
In detail, it comprises three indispensable components, i.e., volume-level query embedding, the feature lifting module, and the MV volume aggregation module.
Note that, in the following discussion, the subscripts $\rm{l}$ and $\rm{n}$ may be omitted to keep simplicity.

{\bf Volume Query Representation.}
The scene is discretized into voxels with shape $\mathrm{Z \times Y \times X}$, 
and assigned with a group of volume-shaped learnable parameters $\mathcal{Q} \in \mathbb{R}^{\mathrm{C\times Z \times Y \times X}}$ as the queries, 
where $\mathrm{X,Y,Z}$ are the spatial shape of the volume and $\mathrm{C}$ governs the hidden dimension.
Specifically, the query $\mathcal{Q}^\mathrm{p} \in \mathbb{R}^{\mathrm{C}}$ located at $\rm{p} = \rm{(d, h, w)}$ is responsible for the corresponding voxel in the volume, 
each voxel in the volume corresponds to a real-world size with $\mathrm{s}$ meter, 
and the center of the volume is aligned with the origin point in the 3D scene coordinate system.
In comparison to existing approaches \cite{zhang2019wide,zheng2021learning,zhang20203d} that utilize Inverse Perspective Mapping (IPM) \cite{bertozz1998stereo} strategy or Spatial Transformer Networks (STN) \cite{jaderberg2015spatial} module for deriving the volume representation of the scene, 
the query-based attention paradigm provides greater scalability, does not rely on the flat ground assumption, and requires less hyperparameter tuning, making it a more practical approach.

{\bf Volume-Level Camera Embedding.}
Existing MVP approaches \cite{tong2023scene,huang2023tri,yang2023bevformer,li2022bevformer,zhou2022cross,liao2022maptr} consider $\mathcal{Q}$ as the universal query and do not take into account the camera's intrinsic and extrinsic parameters, which is comprehensible since MVP task comprises a fixed and stable camera layout \cite{caesar2020nuscenes,sun2020scalability}, e.g., the golden MVP benchmark nuScene \cite{caesar2020nuscenes} comprises 6 stable surrounding cameras to provide 360$^\circ$ FOV.
Nevertheless, the view-agnostic global query $\mathcal{Q}$  is inadequate in addressing the complexities
posed by challenging surveillance environments, where significantly diverse dynamic camera layouts are inevitable
in such scenarios. 
For instance, the CVCS benchmark \cite{zhang2021cross} consists of images from tens of thousands of camera views.
To tackle this problem, we develop a crucial volume-level camera embedding module that utilizes a similar strategy as the image-level camera embedding, which encodes the camera parameters $\{\mathrm{M,K}\}$ and volume-level augmentation matric $\rm A_v \in \mathbb{R}^{2\times 3}$ into the versatile $\mathcal{Q}$ to create the view-dependent volume query with
\begin{align}
\small
  \mathrm{E}_\mathrm{v}
  \Big(\mathcal{Q},\{\mathrm{M,K,A_v}\}\Big)=\mathcal{Q}\otimes \phi_\mathrm{v}\Big([\xi,\mathcal{P}_\mathrm{v}]\Big)  \label{eq:vox_embed},
\end{align}
where $\phi_\mathrm{v}:\mathbb{R}^{30}\rightarrow\mathbb{R}^{\mathrm{C}}$ fuses the positional encoding with the camera parameters, $\mathcal{P}_\mathrm{v} \in \mathbb{R}^{3\times \mathrm{Z\times Y \times X}}$ stands for the positional encoding with $\mathcal{P}_\mathrm{v}(\cdot, \rm{d, h, w})=[\nicefrac{\rm{d}}{\mathrm{Z}}, \nicefrac{\rm{h}}{\mathrm{Y}}, \nicefrac{\rm{w}}{\mathrm{X}}]^\mathrm{T}$, and $\xi$ takes similar definition as in Equation \ref{eq:camera_embed}.

{\bf Feature Lifting Module.} 
Considering the need for supporting dynamic camera layouts and the complex deployment environment in the MVC task, 
employing deformable attention \cite{zhu2020deformable,li2022bevformer} would be more suitable than relying on the IPM paradigm for constructing the feature lifting module. 
Employing deform attention to lift image feature to 3D scene space has been widely used in
multi-view tasks \cite{li2022bevformer,jiang2023polarformer,wang2023frustumformer,li2023fb}.
In this work, the CountFormer also adopts this method but replaces the global versatile query with a view-dependent query to address the complexities in surveillance environments.
Mathematically, given the camera dependent query $\mathcal{Q}$ encoded with Equation \ref{eq:vox_embed}, and the corresponding image-level feature $\mathcal{F}$, 
for each voxel query  $\mathcal{Q}^\mathrm{p}$ located at $\mathrm{p=(d,h,w)}$, 
one may first recover the corresponding 3D location $\mathrm{(x^\prime,y^\prime,z^\prime)}$ {w.r.t.} the scene with
\begin{align}
  \label{eq:query_to_xyz}
  \mathrm{x^\prime=s\cdot(w-\nicefrac{\mathrm{X}}{2});~~y^\prime=s\cdot(h-\nicefrac{\mathrm{Y}}{2});~~z^\prime=s\cdot(d-\nicefrac{\mathrm{Z}}{2})},
\end{align}
then obtain the homogeneous reference point $\mathrm{p=(u,v)}$  by projecting the 3D scene-level point $\mathrm{(x^\prime, y^\prime, z^\prime)}$ according to the intrinsic $\mathrm{K}$ and extrinsic $\mathrm{M}$ as
\begin{align}
  \mathrm{[u,v,1]^\mathrm{T}=\mathrm{K}\cdot \mathrm{M}\cdot[x^\prime,y^\prime,z^\prime,1]^T},
\end{align}
and finally, perform the sophisticated deformable attention mechanism at the projected location $\rm (u,v)$.

{\bf Multi-View Volume Aggregation.}
The MV volume aggregation module aggregates the MV volume representations to produce a comprehensive volume of the scene.
Existing fusion approaches can only deal with stable and fixed camera layouts \cite{li2022bevformer,zhang2019wide,zhang2022wide,zhang20223d,zhang20203d}, 
or are compromised in conducting the fusion process by only considering the geometric position of the cameras and ignoring the critical semantic features \cite{zhang2022wide,zhang2019wide,zhang2021cross}, 
or assume that different views contribute equally to the scene representation \cite{li2022bevformer,huang2021bevdet,li2023bevdepth}.
In summary, all existing approaches lack the necessary scalability to effectively handle dynamic camera layouts.
In this work, a compact yet effective attention mechanism is devised to aggregate the MV volume representations $\{\mathcal{V}_{\rm n}\}_{\rm n \le {\mathrm{N}}}$ with
\begin{align}
\small
  \label{eq:blend}
  \mathcal{V}=\sum\nolimits_{\rm n}\mathcal{W}_{\rm n} \odot \mathcal{V}_{\rm n},
\end{align}
where $\mathcal{W}_{\rm n} \in \mathbb{R}^{1\times \mathrm{Z\times Y \times X}}$ denotes the attention weight, $\odot$ refers the element-wise product operation, and $\mathcal{V}_{\rm n} \in \mathbb{R}^{\mathrm{C \times Z \times Y \times X}}$ is the volume feature w.r.t. the $\rm n_{\mathrm{th}}$ view.
Moreover, instead of employing the computationally heavily vanilla self-attention \cite{vaswani2017attention}, we tend to estimate the weight of the ${\rm p}_{\mathrm{th}}$ voxel in the $\rm n_{\mathrm{th}}$ view $\rm \mathcal{W}_n^p$ much cheaper,
\begin{align}
  \mathcal{W}_{\rm n}^{\rm p}=
  \begin{cases}
    \nicefrac{ \exp \{\phi(\mathcal{Q}_{\rm n}^{\rm p})\}}{\sum_{\rm k \in \mathbf{S}}  \exp \{\phi(\mathcal{Q}_{\rm k}^{\rm p})\}}, & \mathrm{if}~{\rm n} \in \mathbf{S} \\
    0, &\mathrm{others}
  \end{cases}
\end{align}
where $\mathbf{S}$ represents the whole camera-views that voxel $\rm p$ hits, 
$\phi:$ $~\mathbb{R}^{\mathrm{C}} \rightarrow \mathbb{R}_+$ measures the importance of voxel, 
and $\mathcal{Q}_{\rm n}$ referes to the volume query with camera-encoding w.r.t. the $\rm n_{\mathrm{th}}$ view.
\textcolor{black}{It is worth noting that the aggregation strategy possesses significant advantages over previous methods.
For instance, it is independent of the permutation of the camera views and thus is capable of handling dynamic camera layouts.
Besides, the blending weights $\rm \mathcal{W}_n^p$ inherently encode the semantic contextual and geometric position, making it more practical than previous fusion strategies \cite{zhang2021cross,li2022bevformer,huang2021bevdet}, especially dealing with occlusion.
 It shall point out that the MVP tasks necessitate the multi-cameras to provide 360$^\circ$ FOV comprehensive perception of the scene, where marginal overlapping exists between different cameras \cite{caesar2020nuscenes,sun2020scalability}, 
making the cross-view fusion strategy not necessitates elaborated design \cite{li2022bevformer,huang2021bevdet,wei2023surroundocc}.
In contrast, MVC tasks heavily rely on the overlapping between different camera views to address the ambiguity caused by occlusion and the scale variation resulting from perspective projection \cite{zhang2019wide,zhang2021cross}. 
As a result, the multi-view fusion strategy needs to be carefully designed to effectively integrate information from multiple views.}

\subsection{Density Predictor}
As 3D density estimation requires more low-level features to enable the network to learn fine-grained density, we incorporate the 3D FPN \cite{lin2017feature} to perform multi-scale volume feature fusion.
Formally, given multi-scale 3D volumes $\{\mathcal{V}^{\rm l}\}_{\rm l \le \mathrm{L}}$, we upsample ${\rm l-1}_\mathrm{th}$ level 3D density features $\mathcal{X}_{\rm l-1}$ with 3D deconvolution layer and fuse it with $\rm l_{\rm th}$ volume representation $\mathcal{V}_{\rm l}$ as
\begin{align}
\small
  \mathcal{X}_{\rm l} = \mathcal{V}_{\rm l} + \mathrm{DeConv}(\mathcal{X}_{\rm l-1}),
\end{align}
we apply 3D convolution layer on $\mathcal{X}_{\rm l}$ to estimate the 3D density map $\mathcal{G}_{\rm l}$,
and supervise the training procedure with $\mathrm{L}_2 ~ \mathrm{Norm}$.
Considering that the high-resolution prediction remains more important, loss weight $\alpha_{\rm l}$ is employed to balance the training losses among various resolutions.
Besides, as shown in Figure \ref{fig:arch}, we add the 2D density estimation task as an intermediate supervision to accelerate the training process, which may aid in gradient propagation and improve the overall training procedure. This design is reasonable because the complicated CountFormer architecture makes the gradient feedback from the 3D density supervision too long, while the intermediate 2D density estimation task elegantly improves the gradient feedback instead.
To this end, the training objective of the CountFormer is
\begin{align}
\small
  \mathcal{L}=\lambda \left|\mathcal{H}-\Bar{\mathcal{H}}\right|_2 + \sum\nolimits_{\rm l} \alpha_{\rm l}\left| \mathcal{G}_{\rm l}-\Bar{\mathcal{G}_{\rm l}}\right|_2,
\end{align}
where $\left|\cdot\right|_2$ is the $\mathrm{L}_2 ~ \mathrm{Norm}$, $\lambda$ trades between the 2D density supervision and 3D density estimation, 
$\mathcal{H}$ and $\mathcal{G}_{\rm l}$ denote the GT 2D and 3D density, and $\Bar{\cdot}$ is the corresponding prediction.
Compared to the complicated loss design strategy adopted in \cite{zhang2019wide,zhang2022wide,zhang20203d,zhang20223d,zhang2022calibration}, the training objective in CountFormer is much simpler yet more straightforward to tune.

\section{Experiments}
\subsection{Experiment Settings}
\vspace{-3pt}
We leverage all existing MVC datasets to evaluate the effectiveness of the CountFormer, 
including CityStreet \cite{zhang2019wide}, PETS2009 \cite{ferryman2009pets2009}, DukeMTMC \cite{ristani2016performance}, and CVCS \cite{zhang2021cross}.
To make a fair comparison, following conventional works \cite{zhang2019wide,zhang2022wide,zhang2022calibration,zhang20203d,zhang20223d}, we employ the mean absolute error ($\mathrm{MAE} \downarrow$) and normalized mean absolute error ($\mathrm{NAE} \downarrow$) as the evaluation criteria to quantify the counting performances on both image-level and scene-level.
Our code and model is available at \url{https://github.com/MandyMo/ECCV_Countformer} for research purpose.

\subsection{Qualitative Experiments}
\begin{figure*}[h]
  \centering
  \includegraphics[width=0.85\textwidth]{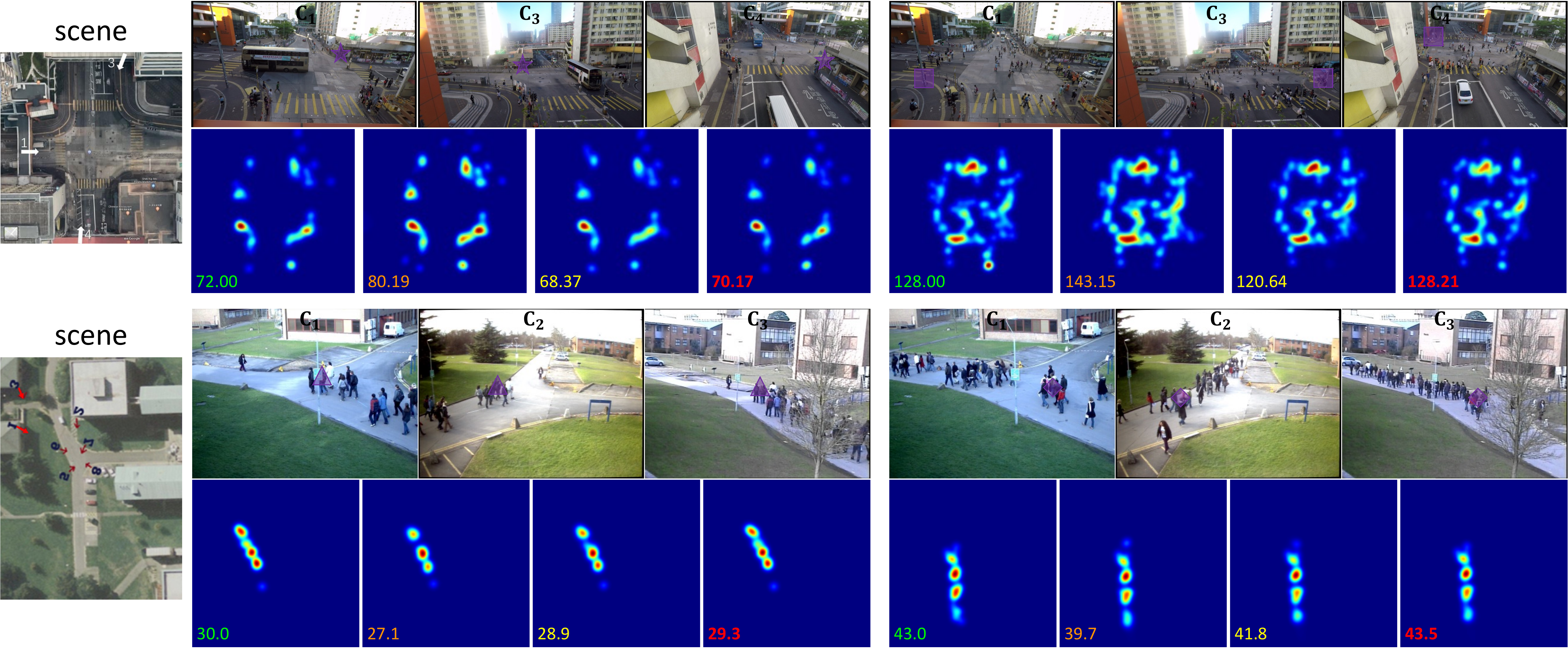}
  \vspace{-5pt}
  \caption{{\bf Qualitative Results.} 
  The figure exhibits several typical scenarios on the CityStreet (with 3 views) and PETS2009 (with 3 views) datasets, including occlusion and congested crowds.   
  For each sample, the multi-view images, the {\color{green} ground truth} scene-level density and estimated density from {\color{orange}CVCS method}\cite{zhang2021cross}, {\color{yellow}3D Counting approach}\cite{zhang20203d}, and {\color{red}the CountFormer} are presented in the bird's eye view, respectively.}
  \label{fig:visualize_citystreet_pets2009}
\end{figure*}

\vspace{-3pt}
To better demonstrate the robustness of CountFormer in the challenging scenarios, we draw some representative samples from CityStreet and PETS2009 testing sets, as illustrated in Figure \ref{fig:visualize_citystreet_pets2009}, i.e., in the presence of occlusion and the congested crowds.
Experiments demonstrate the capability of CountFormer in dealing with occlusion.
Specifically, the CVCS approach\cite{zhang2021cross} encounters difficulties in accurately reconstructing the density in spatial regions obstructed by objects such as buses or trees.
By employing the MV volume aggregation module, our CountFormer can alleviate this issue and make a reasonable estimation, because the aggregation module is capable of dynamic blend voxel features from all views.
Moreover, both the CVCS \cite{zhang2021cross} and the 3D Counting \cite{zhang20203d} demonstrate limited efficacy in managing densely populated crowds.
As expected, CountFormer effectively tackled these challenges by utilizing the feature lifting module to transform image-level features to 3D volume for each view and adopting the MV volume aggregation module to attentively fuse suitable features for each voxel.

\begin{figure*}[!]
  \centering
  \includegraphics[width=0.85\textwidth]{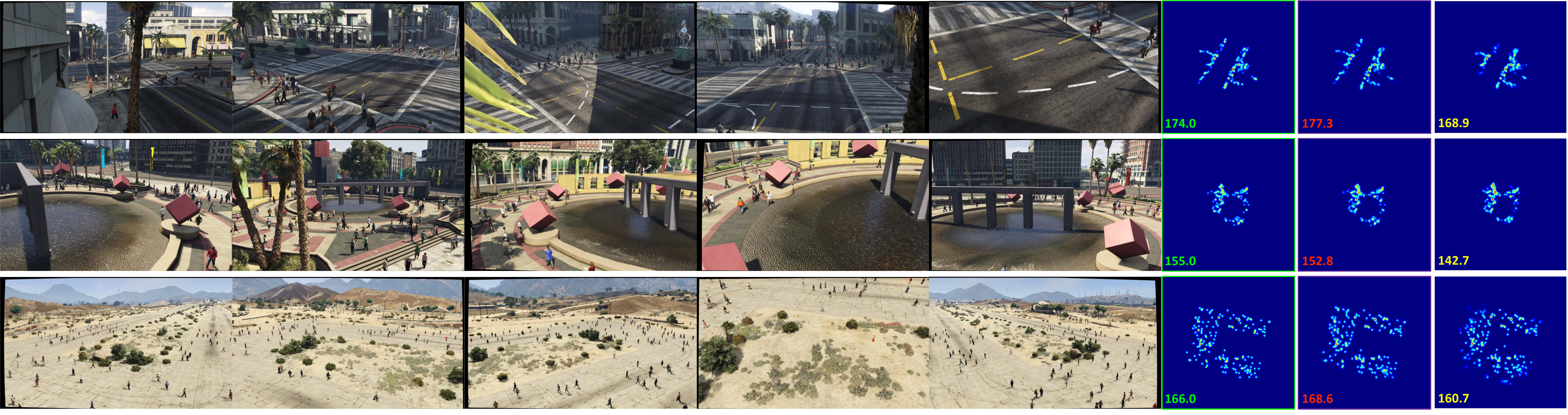}
  \caption{{\bf Qualitative Results.} 
  The figure visualizes 3 challenging scenarios on the CVCS benchmark.
  Regarding each sample, the visualization includes the multi-view images (with 5 views), {\color{green}ground truth} density, density obtained {\color{red} with the MV volume aggregation module}, and density estimated {\color{yellow}without this module}.
  }
  \label{fig:visualize_cscv}
\end{figure*}

\textcolor{black}{To gain a better understanding of the effectiveness of the feature aggregation module in CountFormer, 
we select a 3D reference point for each sample and report the fusion weight of each point across all views (the projected reference points are marked as \textcolor{purple(html/css)}{$\bigstar$}, \textcolor{purple(html/css)}{$\square$}, \textcolor{purple(html/css)}{$\vartriangle$}, and \textcolor{purple(html/css)}{$\lozenge$} in Figure \ref{fig:visualize_citystreet_pets2009}).
As Table \ref{tb:atten_weight} reports the attention weights of each 3D reference point, 
CountFormer assigns smaller weights to occluded views because the MV volume aggregation module jointly consider the geometric information and semantic feature of each view, while the CVCS approach focuses on geometric camera position only (see \textcolor{purple(html/css)}{$\bigstar$}, \textcolor{purple(html/css)}{$\square$}, and \textcolor{purple(html/css)}{$\vartriangle$} ).
Additionally, CountFormer also automatically assigns appropriate weights to views with congested crowds, enhancing the count accuracy (see \textcolor{purple(html/css)}{$\lozenge$}).
}

\begin{table}[htbp]
  \centering
  \begin{subtable}{0.5\linewidth}
    \centering
    \begin{tabular}{c|l|ccc}
      \toprule
      \multicolumn{1}{c}{\multirow{2}{*}{\bf Sample}}  & \multicolumn{1}{|c|}{\multirow{2}{*}{\bf Method}} & \multicolumn{3}{c}{\bf Atten-Weight} \\\cline{3-5}
      & & \bf C$_1$ & \bf C$_3$ & \bf C$_4$ \\
      \hline
      \multirow{2}{*}{\textcolor{purple(html/css)}{$\bigstar$}} 
      & CVCS& 0.15 & 0.23 & 0.62 \\
      & CountFormer & 0.40 & 0.51 & 0.09 \\
      \hline
      \multirow{2}{*}{\textcolor{purple(html/css)}{$\square$}} 
      & CVCS & 0.67 & 0.22 & 0.11 \\
      & CountFormer & 0.18 & 0.33 & 0.49 \\
      \bottomrule
    \end{tabular}
    \caption{Attention Weights on CityStreet.}
  \end{subtable}%
  \hfill
  \begin{subtable}{0.5\linewidth}
    \centering
    \begin{tabular}{c|l|ccc}
      \toprule
      \multicolumn{1}{c}{\multirow{2}{*}{\bf Sample}}  & \multicolumn{1}{|c|}{\multirow{2}{*}{\bf Method}} & \multicolumn{3}{c}{\bf Atten-Weight} \\\cline{3-5}
      & & \bf C$_1$ & \bf C$_2$ & \bf C$_3$ \\
      \hline
      \multirow{2}{*}{\textcolor{purple(html/css)}{$\vartriangle$}} 
      & CVCS & ~0.38~ & ~0.41~ & ~0.21~ \\
      & CountFormer & 0.19 & 0.45 & 0.36 \\
      \hline
      \multirow{2}{*}{\textcolor{purple(html/css)}{$\lozenge$}}  
      & CVCS & 0.35 & 0.43 & 0.28 \\
      & CountFormer & 0.45 & 0.47 & 0.08 \\
      \bottomrule
    \end{tabular}
    \caption{Attention Weights on PETS2009.}
  \end{subtable}
  \caption{{\bf Quantitative Analysis of Attention Weights.} 
  The table presents the attention weights according to the MV features on CityStreet and PETS2009, 
  where the 3D reference points \textcolor{purple(html/css)}{$\bigstar$} , \textcolor{purple(html/css)}{$\square$}, \textcolor{purple(html/css)}{$\vartriangle$}, and \textcolor{purple(html/css)}{$\lozenge$} according each view are marked on Figure \ref{fig:visualize_citystreet_pets2009}.
  }
  \vspace{-15pt}
  \label{tb:atten_weight}
\end{table}

\textcolor{black}{Figure \ref{fig:visualize_cscv} depicts the density maps w/ and w/o the volume aggregation module on the challenging CVCS benchmark, 
one may observe that, 
w/o the aggregation module, 
it tends to undercount in spatial regions that are occluded in some views (2$_{\rm nd}$ sample), 
the model fails to conduct exact counting at the present of exaggerated views (1$_{\rm st}$ sample), 
and the CountFormer may struggle to predict sharp density maps with large-scale variation (3$_{\rm rd}$ sample).
However, w/ the aggregate module, CounFormer is capable of dealing with these challenges, highlighting this module's ability to effectively fuse MV volume features, especially in situations with significant scale variations or occlusions.
}

\begin{figure}[htbp]
  \centering
  \begin{subfigure}[b]{0.4\textwidth}
    \includegraphics[width=\textwidth]{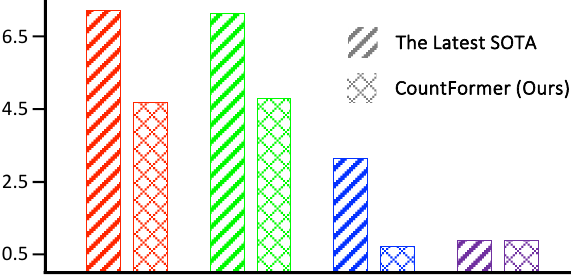}
    \caption{Histogram Comparison}
  \end{subfigure}
  \hfill
  \begin{subtable}[b]{0.5\textwidth}
    \begin{tabular}{l|c|c}
      \toprule
      ~~~\bf Dataset~ & ~~~\bf SOTA~~ & ~\bf Ours~~ \\
      \midrule
      \textcolor{red}{CityStreet} \cite{zhang2019wide} & 6.98 \cite{zhang2022wide} & 4.72 \\
      \textcolor{green}{CVCS} \cite{zhang2021cross}& 7.22 \cite{zhang2021cross} & 4.79 \\
      \textcolor{blue}{PETS2009} \cite{ferryman2009pets2009} & 3.08 \cite{zheng2021learning} & 0.74 \\
      \textcolor{purple(html/css)}{DukeMTMC}\cite{ristani2016performance} & 0.87 \cite{zheng2021learning} & 0.88 \\
      \bottomrule
    \end{tabular}
    \caption{Quantitative Comparison}
  \end{subtable}
  \caption{{\bf{Comparisons with state-of-the-art (SOTA) methods.}} 
  The figure presents the comparisons between \cite{zhang2019wide,zhang2022wide,zhang20203d,zhang20223d,zhang2021cross,zheng2021learning,qiu2019cross,zhang2022calibration} and our CountFormer, where the mean absolute error~(MAE~$\downarrow
  $) is used to evaluate the performance on the \textcolor{red}{CityStreet dataset}\cite{zhang2019wide}, \textcolor{green}{CVCS dataset} \cite{zhang2021cross}, \textcolor{blue}{PETS2009 dataset}\cite{ferryman2009pets2009}, and \textcolor{purple(html/css)}{DukeMTMC dataset}\cite{ristani2016performance}.
  For better visualization, we plot the best performance among \cite{zhang2019wide,zhang2022wide,zhang20203d,zhang20223d,zhang2021cross,zheng2021learning,qiu2019cross,zhang2022calibration} to compare with ours on each dataset.
  }
  \label{fig:compare}
  \vspace{-10pt}
\end{figure}

\subsection{Quantitative Experiments}
\vspace{-3pt}
As Figure \ref{fig:compare} briefly summarizes the comparison of CountFormer against the latest state-of-the-art (SOTA) performances, 
the CountFormer substantially outperforms the SOTA by a large margin on PETS2009 \cite{ferryman2009pets2009}, CityStreet\cite{zhang2019wide}, and CVCS \cite{zhang2021cross} datasets,
and achieves comparable performance on DukeMTMC \cite{ristani2016performance}.

Specifically, on the {\bf DukeMTMC dataset}\cite{ristani2016performance}, our method achieves competitive results among newly launched approaches 
\cite{zhang2019wide,zhang2022wide,zhang20203d,zhang20223d,zhang2021cross,zheng2021learning,zhang2022calibration,qiu2019cross}.
Although the performance seems not overwhelming, it is reasonable considering the saturated performance \cite{zhang2022wide,zhang20223d,zhang2021cross,zheng2021learning}, the inaccurate annotation quality \cite{zhang2019wide}, and the non-overlapping camera view.
In contrast, the {\bf PETS2009 dataset} \cite{ferryman2009pets2009} contains a congested crowd distribution and overlapping views. 
As a result, the CountFormer significantly outperforms all existing approaches \cite{zhang2019wide,zhang2022wide,zhang20203d,zhang20223d,zhang2021cross,zheng2021learning,qiu2019cross,zhang2022calibration} and achieves exceptional performance, reducing the scene-level $\mathrm{MAE}$ by $76.0\%$.
Similarly, a similar trend can be seen in the {\bf CityStreet benchmark} \cite{zhang2019wide}, which contains a larger crowd distribution, severe dynamic occlusions from the environment, and diverse scale variations caused by perspective projection, 
the CountFormer significantly outperforms all current approaches \cite{zhang2019wide,zhang2022wide,zhang20203d,zhang20223d,zhang2021cross,zheng2021learning,qiu2019cross,zhang2022calibration,zhang2022calibration} in terms of $\mathrm{MAE/NAE}$ for both scene-level and single-view level, maintaining a state-of-the-art performance and reducing the scene-level MAE/NAE by nearly half.
On the large-scale and most challenging MVC benchmark {\bf CVCS} \cite{zhang2021cross}, 
CountFormer achieves an impressive $\mathrm{MAE/NAE}$ and sets a new state-of-the-art performance, outperforming the latest 3D counting approach \cite{zhang20223d} by $178\%$.

It is not astonishing when considering the superior architecture of the CountFormer.
Specifically, (1) the Feature Lifting Module employs the attention mechanism to retrieve semantic features from image space, which proves more robust than the IPM \cite{bertozz1998stereo} strategy that previous works adopted \cite{zhang2022wide,zhang20223d} because the IPM assumes the world to be flat on a plane, which seems difficult to be guaranteed. 
Moreover, it remains complicated to consider the height of people in the crowd for the IPM strategy, making the IPM tend to commit misaligned feature transformation;
(2) the MV Volume Aggregation Module is capable of solving the occlusion and scale variation problem by picking features from the appropriate view, yet contemporary works \cite{zhang2021cross} turn to fuse multi-view features based on the distance of the IPM plane to the camera while ignoring the semantic feature themselves.
While distance prior is capable of solving the scale variation dilemma, it seems powerless when dealing with occlusion since occlusion necessitates the semantic features for further validation;
(3) the camera encoding strategy embeds the camera parameters into the volume query and the image features, implicitly allowing the CountFormer to model the camera extrinsic and intrinsic and facilitating the CountFormer to deal with arbitrary camera layouts.

  \begin{table}[t]
    \centering
    \begin{tabular}{c c c  c | c c | c c | c c}
    \toprule[1pt]
    \multicolumn{1}{c}{\multirow{2}{*}{$\mathrm{~L~}$}} &  \multicolumn{1}{c}{\multirow{2}{*}{$\mathrm{~V~}$}} &  \multicolumn{1}{c}{\multirow{2}{*}{$\mathrm{~A~}$}} & \multicolumn{1}{c|}{\multirow{2}{*}{$\mathrm{~I~}$}} &\multicolumn{2}{c|}{PETS2009} & \multicolumn{2}{c|}{CityStreet} & \multicolumn{2}{c}{CVCS}\\\cline{5-10}
    & & & &$\mathrm{MAE}$&$\mathrm{NAE}$&$\mathrm{MAE}$&$\mathrm{NAE}$&$\mathrm{MAE}$&$\mathrm{NAE}$ \\
    \hline
         &      &       &      & ~~2.76~ & ~0.114~ & ~7.12~ & ~0.084~ &~11.3~ & ~0.088~ \\
    \yes &      &       &      & 1.31 & 0.054 & 5.65 & 0.071 & 9.51 & 0.074 \\
    \yes & \yes &       &      & 1.17 & 0.047 & 5.15 & 0.062 & 6.22 & 0.048 \\
    \yes & \yes & \yes  &      & 0.80 & 0.033 & 4.81 & 0.057 & 4.96 & 0.041 \\
    \yes & \yes & \yes  & \yes & 0.74 & 0.030 & 4.72 & 0.058 & 4.79 & 0.039 \\
    \bottomrule[1pt]
    \end{tabular}
    \vspace{2pt}
    \caption{{\bf Ablation Study.} 
    The table presents the ablation results w.r.t. various combinations of the critical components, 
    where {\bf A} adopts the MV Volume Aggregation Module, 
    {\bf \xcancel{A}} simply averages the MV volume for each voxel, 
    {\bf I} and {\bf \xcancel{~I~}} denotes injecting camera-encoding into image-level features or not, 
    {\bf V} and {\bf \xcancel{~V~}} governs whether adopting volume-level camera encoding, 
    {\bf L} refers to the Feature Lifting Module, 
    and {\bf \xcancel{~L~}} naively employs the IPM to transform image features to 3D space.
    }
    \vspace{-20pt}
    \label{tb:abs_comprehensive}
  \end{table}

\subsection{Ablation Study}
\vspace{-3pt}
In this section, various ablation experiments are conducted to comprehend the merit of the CountFormer competently.

Firstly, we evaluate the impact of the critical designs that constitute the CountFormer. 
As Table \ref{tb:abs_comprehensive} summarizes the ablation results, adopting the Feature Lifting Module substantially improves the performances on the PETS2009 and CityStreet datasets because PETS2009 and CityStreet comprise fixed camera layouts, making the Feature Lifting Module adequate to lift image features to volume representation and favoring the volume queries to encode the camera layouts implicitly, overwhelming the boosting of the performances.
\begin{table}
  \centering
  \begin{tabular}{l |c c c |c c |c c|c c}
  \toprule[1pt]
  \multicolumn{1}{c|}{\multirow{2}{*}{Method}} & \multicolumn{1}{|c}{\multirow{2}{*}{$\mathrm{~L~}$}} & \multicolumn{1}{c}{\multirow{2}{*}{$\mathrm{~V~}$}} & \multicolumn{1}{c|}{\multirow{2}{*}{$\mathrm{~A~}$}} &\multicolumn{2}{c|}{PETS2009} & \multicolumn{2}{c|}{CityStreet} & \multicolumn{2}{c}{CVCS}\\\cline{5-10}
  & & & & $\mathrm{MAE}$ & $\mathrm{NAE}$ & $\mathrm{MAE}$ & $\mathrm{NAE}$ & $\mathrm{MAE}$ & $\mathrm{NAE}$ \\
  \hline
  \multirow{4}{*}{3D Counting \cite{zhang20203d}} 
  &      &      &      & ~3.25~ & ~0.136~ & ~7.63~ & ~0.102~ & ~12.8~ & ~0.116~  \\
  &\yes  &      &      & 2.53 & 0.102 & 6.76 & 0.091 & 12.9 & 0.114  \\
  &\yes  &\yes  &      & 2.18 & 0.084 & 6.09 & 0.082 & 8.52 & 0.069 \\
  & \yes & \yes & \yes & 1.29 & 0.051 & 5.54 & 0.074 & 6.21 & 0.046 \\
  \hline
  \multirow{4}{*}{CVCS \cite{zhang2021cross}}
  &      &      &      & 3.81  & 0.142 & 7.43 & 0.101 & 7.27 & 0.061   \\
  &\yes  &      &      & 2.79  & 0.108 & 7.14 & 0.098 & 7.41 & 0.063   \\
  &\yes  & \yes &      & 2.54  & 0.097 & 6.86 & 0.094 & 7.01 & 0.059   \\
  & \yes & \yes &\yes  & 1.72  & 0.063 & 6.17 & 0.080 & 6.76 & 0.054   \\
  \hline
  \multirow{3}{*}{BEVFormer \cite{li2022bevformer}}
  &\yes         &      &      & 3.04  & 0.119 & 7.17 & 0.096 & 9.56 & 0.087   \\
  &\yes  & \yes &      & 2.63  & 0.105 & 6.63 & 0.085 & 7.32 & 0.061   \\
  & \yes & \yes &\yes  & 1.45   & 0.058 & 5.81 & 0.078 & 6.33 & 0.048   \\
  
  \bottomrule[1pt]
  \end{tabular}
  \vspace{2pt}
  \caption{{\bf Ablation Study.}
  The table shows the effects of integrating critical components into existing approaches \cite{zhang20223d,zhang2021cross,li2022bevformer}, 
  where {\bf L}, {\bf V}, {\bf A}, {\bf \xcancel{L}}, {\bf \xcancel{V}}, and {\bf \xcancel{A}}
  shares a similar definition as in Table \ref{tb:abs_comprehensive}.
  }
  \vspace{-18pt}
  \label{tb:abs_cvcs_3d}
\end{table}
Furthermore, it is noteworthy that the integration of the Volume-Level Camera Embedding alongside the MV volume Aggregation Module exhibits a substantial enhancement in performance when applied to the CVCS dataset, which currently stands as one of the most challenging benchmarks for MVC analysis.
We believe the notable enhancement in performance attributed to the Volume-Level Camera Embedding's explicit incorporation of camera extrinsic and intrinsic parameters, as well as the MV Volume Aggregation Module's aptitude for selecting suitable features for each voxel.
Furthermore, the introduction of camera encoding into image features yields marginal performance improvement, 
which is reasonable as the camera encoding complements the attention mechanism within the feature-lifting module.

\begin{table}
    \centering
    \begin{tabular}{c|*{4}{c}}
    \toprule[1pt]
    \multicolumn{1}{c|}{\multirow{2}{*}{\#Train-View}} & \multicolumn{4}{c}{{\#Test-View}} \\\cline{2-5}
    & 3 & 5 & 7 & 9 \\
    \hline
    $\star$            & ~23.7/0.178~ & ~11.0/0.081~ & ~6.24/0.047~ & ~3.88/0.029~ \\
    5                  & 26.8/0.197 & 11.0/0.081 & 8.06/0.059 & 4.52/0.034 \\
    $\mathrm{U}(2,11)$ & 24.0/0.179 & 11.9/0.088 & 6.72/0.052 & 4.02/0.030\\
    \bottomrule[1pt]
    \end{tabular}
    \vspace{5pt}
    \caption{{\bf Ablation Study.}
    The table investigates the robustness of CountFormer against various numbers of camera views on the CVCS dataset \cite{zhang2021cross}, where the number of image views used for training is not necessarily equal to that in the testing stage and the ground-truth count is all the people in the scene.
    Specifically, $\star$ indicates that the \#Train-view equals the \#Test-view, 
    $5$ means the CountFormer is trained with \#Train-View=5 and evaluated with various \#Test-View,
    and $\mathrm{U}(\rm a,b)$ denotes the \#Train-view uniformly sampled from $\rm a$ to $\rm b$ in each iteration.
    }
    \vspace{-15pt}
    \label{tb:ab_cscv_num_cam}
  \end{table}

Table \ref{tb:abs_cvcs_3d} presents the effectiveness of the components when integrated with established MVC methodologies \cite{zhang20203d,zhang2021cross} and MVP approach\cite{li2022bevformer}.
We may observe that the volume-level camera embedding, the MV volume aggregation module, and the feature lifting module can also substantially improve the performances of existing MVC methods \cite{zhang2021cross,zhang20203d}.
Additionally, experimental results on the challenging CVCS dataset demonstrate that the MV volume aggregation and camera-level volume embedding are indispensable in dealing with arbitrary dynamic camera layouts.
\textcolor{black}{Moreover, it is infeasible to naively adopt the established MVP architecture BEVFormer \cite{li2022bevformer} without considering the specific challenges of the MVC settings.
Fortunately, equipped with the proposed components, e.g., volume-level camera embedding and MV volume aggregation module, the BEVFormer achieves promising performances. 
}

    \begin{table}
    \centering
    \begin{tabular}{c c |c c |c c}
    \toprule[1pt]
    \multicolumn{1}{c}{\multirow{2}{*}{\#Train-View}} & \multicolumn{1}{c|}{\multirow{2}{*}{\#Test-View}} &\multicolumn{2}{c|}{PETS2009} & \multicolumn{2}{c}{CityStreet} \\\cline{3-6}
    & &$\mathrm{MAE}$&$\mathrm{NAE}$&$\mathrm{MAE}$&$\mathrm{NAE}$  \\
    \hline
    \multirow{3}{*}{$\star$}
     & ~1~ & ~5.43~ & ~0.217~ & ~7.75~ & ~0.091  \strut\\
     & 2 & 2.07 & 0.082 & 6.03 & 0.071  \strut\\
     & 3 & 0.74 & 0.030 & 4.72 & 0.058  \strut\\
    \hline
    \multirow{3}{*}{3}
     & 1 & 7.05 & 0.267 & 8.42 & 0.100  \strut\\
     & 2 & 3.46 & 0.139 & 6.71 & 0.080  \strut\\
     & 3 & 0.74 & 0.030 & 4.72 & 0.058  \strut\\
    \hline
    \multirow{3}{*}{$\mathrm{U}(1,3)$}
     & 1 & 6.01 & 0.257 & 7.91 & 0.098  \strut\\
     & 2 & 2.56 & 0.991 & 6.29 & 0.074  \strut\\
     & 3 & 0.89 & 0.035 & 5.16 & 0.063  \strut\\
    \bottomrule[1pt]
    \end{tabular}
    \vspace{2pt}
    \caption{{\bf Ablation Study.}
    The table demonstrates the robustness of CountFormer with various numbers of cameras on the PETS2009 \cite{ferryman2009pets2009} and CityStreet \cite{zhang2019wide} datasets,
    where $\star$ and $\mathrm{U}(\cdot, \cdot)$ take similar definition as in Table \ref{tb:ab_cscv_num_cam}.
    }
    \label{tb:ab_cscv_num_cam_pets2009_citystreet}
    \vspace{-15pt}
  \end{table}

Moreover, in challenging surveillance environments, it is inevitable in practical situations that some cameras stand down,
necessitating the capability of dealing with dynamic camera layouts, e.g., various numbers of camera views and cameras with significantly different views.
As Table \ref{tb:ab_cscv_num_cam} and Table \ref{tb:ab_cscv_num_cam_pets2009_citystreet}
demonstrate the performances of CountFormer with dynamic arbitrary camera layouts, 
CountFormer achieves encouraging performances even when some cameras are deactivated during the inference stage.
By incorporating dynamic view selection during the training stage, CountFormer achieves comparable performance to models trained with a predefined number of views. This robustness in handling dynamic camera layouts contributes to the effectiveness of CountFormer in challenging surveillance scenarios.

\vspace{-6pt}
\section{Conclusion and Discussion}
\vspace{-2pt}
3D multi-view counting (MVC) is a challenging research area with many potential real-world applications. Nevertheless, there is currently no method that can solve the 3D MVC problem with arbitrary dynamic camera layouts. 
To address this gap, we developed a concise multi-view learning framework, CountFormer.
Experimental results demonstrated that CountFormer is capable of handling challenging scenarios that single-view counting approaches struggle with. 
Quantitatively, CountFormer substantially outperforms all existing MVC approaches and achieves state-of-the-art performance on most MVC benchmarks.
We believe that CountFormer can provide valuable insights for further research on MVC in real-world scenarios.
In addition, the training process of CountFormer requires labor-intensive 3D annotations of the head point, posing challenges for deployment in real-world scenarios. 
However, labeling 2D head points in image space has proven to be more efficient. Therefore, it deserves to explore methods to leverage 2D annotations for training CountFormer in the future.
Furthermore, Resource consumption is a significant problem that requires careful consideration in real-world scenarios. 
Employing more efficient attention mechanisms, using channel pruning methods, as well as adopting the quantization techniques to accelerate the inference speed is a promising work.
%
{\small
\bibliographystyle{ieee_fullname}
\bibliography{egbib}
}

\end{document}